# Single Object Tracking: A Survey of Methods, Datasets, and Evaluation Metrics

Zahra Soleimanitaleb[a], Mohammad Ali Keyvanrad[a, ∗]

[a] *Faculty of Electrical & Computer Engineering Malek Ashtar University of Technology, Iran*

## ABSTRACT

Object tracking is one of the foremost assignments in computer vision that has numerous commonsense applications such as traffic monitoring, robotics, autonomous vehicle tracking, and so on. Different researches have been tried later a long time, but since of diverse challenges such as occlusion, illumination variations, fast motion, etc. researches in this area continues. In this paper, different strategies of the following objects are inspected and a comprehensive classification is displayed that classified the following strategies into four fundamental categories of feature-based, segmentation-based, estimation-based, and learning-based methods that each of which has its claim sub-categories. The most center of this paper is on learning-based strategies, which are classified into three categories of generative strategies, discriminative strategies, and reinforcement learning. One of the sub-categories of the discriminative show is deep learning. Since of high-performance, deep learning has as of late been exceptionally much consider. Finally, the different datasets and the evaluation methods that are most commonly used will be introduced.

## 1 Introduction

Object tracking is one of the foremost assignments in computer vision that attempts to detect and track objects in image sequences. Object tracking has various applications. Object tracking applicable in areas such as traffic monitoring (e.g. monitoring of traffic flow (Tian et al., 2011) and detection of traffic accidents (Tai et al., 2004)), robotics (e.g. ASIMO humanoid robot (Sakagami et al., 2002)), autonomous vehicle tracking (e.g. path-tracking (Brown et al., 2017; Laurense et al., 2017; Menze and Geiger, 2015)), medical diagnosis systems (e.g. tracking of ventricular wall (Colchester and Hawkes, 1991) and medical instruments control (Tsai et al., 2018)), and activity recognition (e.g. learning activity patterns (Stauffer and Grimson, 2000) and human activity recognition (Bodor et al., 2003)).

There are many challenges in object tracking that have led to ongoing research in this area. Some of these challenges in the OTB[1] dataset (Wu et al., 2013) are presented in Table ۱.

As shown in Fig. 1, there are two types of object tracking: single object tracking and multi object tracking.

Single object tracking only track an individual target during a video, and the target specifies in the first frame and must be detected and tracked in the next frames of the video. Single object trackers should be able to track whatever object they are given, even an object on which no available classification model was trained.

In multi object tracking there are multiple objects to track. In this type of tracking, the tracker must first determine the number of objects in each frame and then keep track of each object's identity from frame to frame.

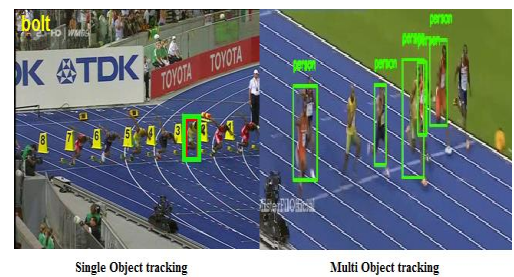

**Fig. 1.** single object and multi object tracking

The aspects of tracking have serious challenges and there are some expectations from the tracking system, so when designing each tracking algorithm, we should try to address those expectations as much as possible. Some of these features are listed below.

**Robustness**: Robustness means that the tracking system can track the target even in complicated conditions such as background clutters, occlusion and illumination variation.

**Adaptability**: In addition to the environment changes, the target is to changes, such as the complex and sudden movement of the target. In order to solve this problem, the tracking system must be able to detect and track the current apparent characteristics of the target.

**Real-time processing of information**: A system that deals with image sequences should have high processing speeds.

So, there is a need to implement a high-performance algorithm.

[1] Object Tracking Benchmark

**Table ١: Some of the object tracking challenges**

| Challenge | Describe | Example |
|---|---|---|
| Illumination Variation | The illumination in the target region is significantly changed. | 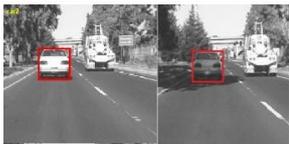 |
| Background Clutters | The background near the target has a similar color or texture as the target. | 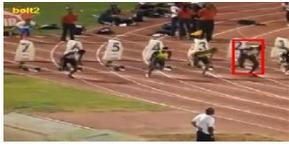 |
| Low Resolution | The number of pixels inside the ground-truth bounding box is low. | 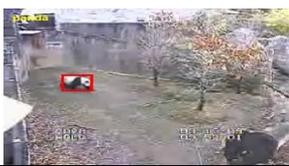 |
| Scale Variation | The ratio of the bounding boxes of the first frame and the current frame is out of the range. | 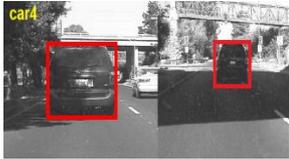 |
| Occlusion | The target is partially or fully occluded. | 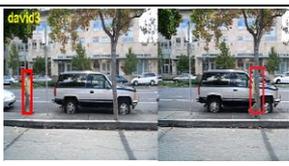 |
| Change the target position | During the movement, the target may be rotated, deformed, and so on. | 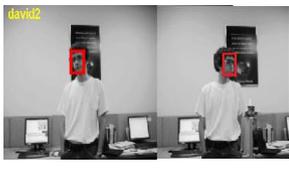 |
| Fast Motion | The motion of the ground truth is large. | 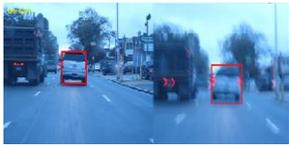 |

## 2 Classification of the object tracking methods

Object tracking methods have different categories, for example, Fiaz et al. have a comprehensive study of tracking methods that categorize tracking methods into two groups of methods based on the correlation filter and the noncorrelation filter (Fiaz et al., 2018). Li et al. have reviewed and compared deep learning methods of object tracking (Li et al., 2018), Verma has reviewed methods of object detecting and tracking and categorized tracking methods into five categories of feature-based, segmentation-based, estimation-based, appearance-based and learning-based methods (Verma, 2017).

In this paper, the last grouping is used and as shown in Fig. 2, a comprehensive classification of single object tracking methods is provided (Soleimanitaleb et al., 2019) and details of each one are given below.

### 2.1 Feature-Based Methods

This method is one of the simple ways of object tracking. To track objects, features, such as color, texture, optical flow, etc., are extracted first. These extracted features must be unique so that objects can be easily distinguishable in the feature space. Once the features are extracted, then the next step is to find the most similar object in the next frame, using those features, by exploiting some similarity criterion.

One of the problems with these methods is at the extraction step because the unique, exact and reliable features of the object should be extracted so that it can distinguish the target from other objects. Here are some of the features that are used for object tracking.

#### 2.1.1 Color

The color features can show the appearance of the object. This feature can be used in different ways, one of the most common methods to use this feature is a color histogram.

The color histogram shows the distribution of colors in an image. in fact, shows the different colors and the number of pixels of each color in the image. The disadvantage of color histograms is that the representation just depends on the color of the object and ignores the shape and texture of the object, so two different objects may have the same histogram.

Some papers have used the color histogram for tracking (Fotouhi et al., 2011; Li and Zheng, 2004; Zhao, 2007).

Zhao (Zhao, 2007) used histograms for tracking; Fig. 3 shows the flowchart of the tracking using this method.

This paper presents a comprehensive classification of object tracking methods and focuses more on learning-based methods.

In section 2, this classification is examined and definitions of different methods and examples of them are given. In section 3 lists the common datasets that used in object tracking are given, and in section **4**. common evaluation methods of object tracking are given.

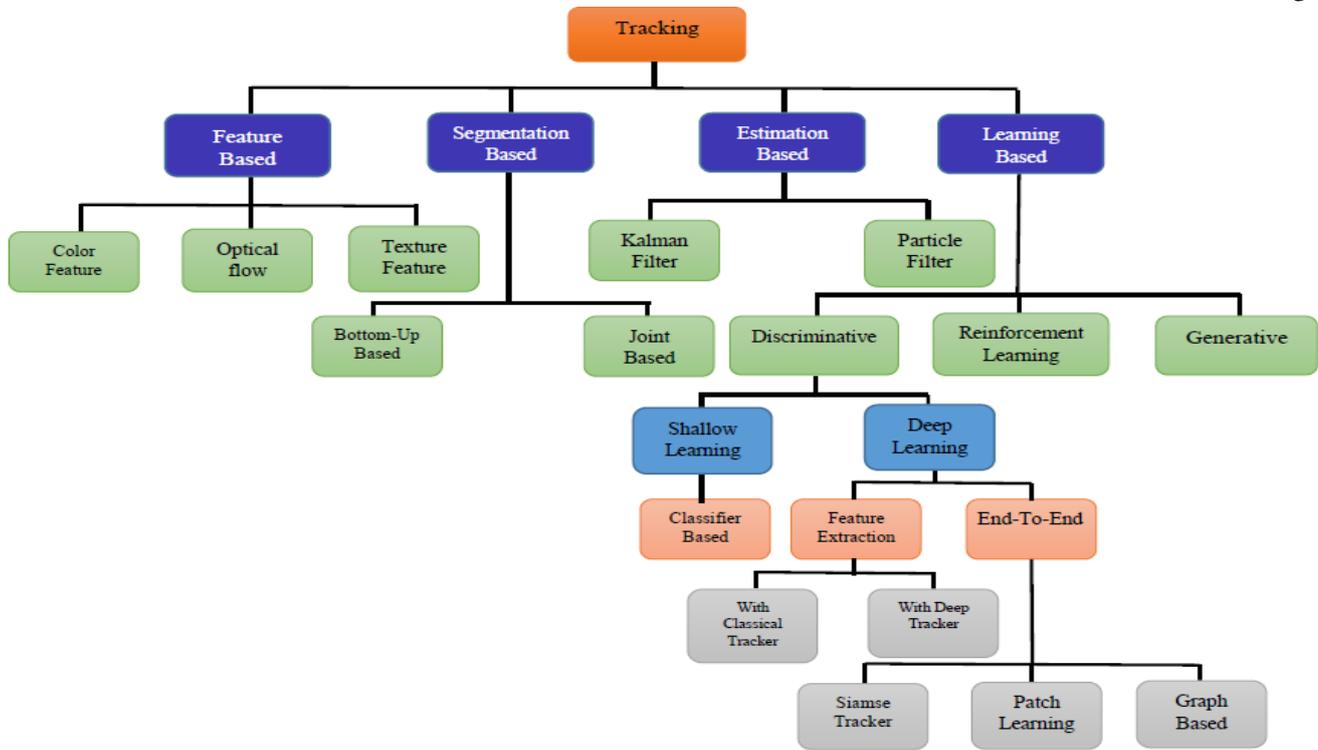

**Fig. 2.** Classification of object tracking methods

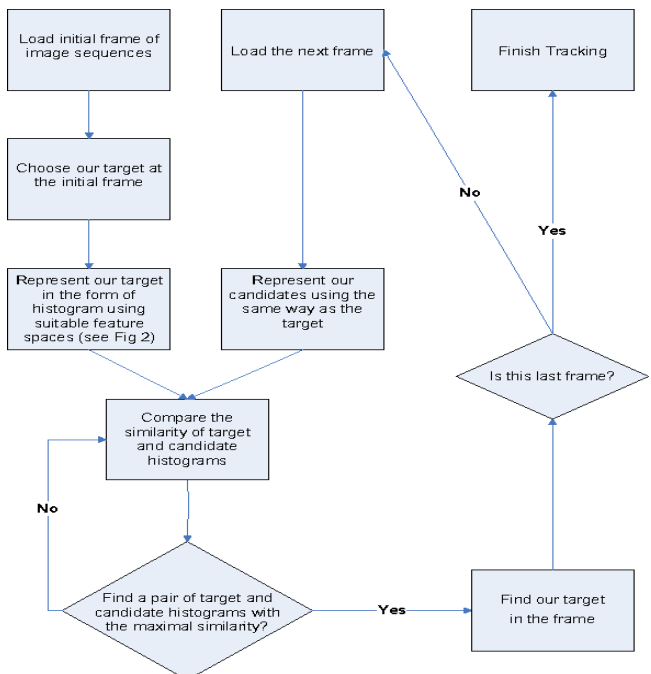

**Fig. 3.** Flowchart of histogram-based object tracking in the gradient feature space (Zhao, 2007)

The tracking is done in such a way that in the first frame the target is marked with a bounding box; then, as seen in Fig. 4, this target is displayed using the histogram in the appropriate feature space. After entering the next frame, the target should be searched in this frame. In the new frame, the method introduces all possible candidates; then, a comparison is made to find a pair with the maximum similarity between the candidate and target histograms.

After finding the target, the next frame is inserted and the same steps are repeated until the last frame is reached.

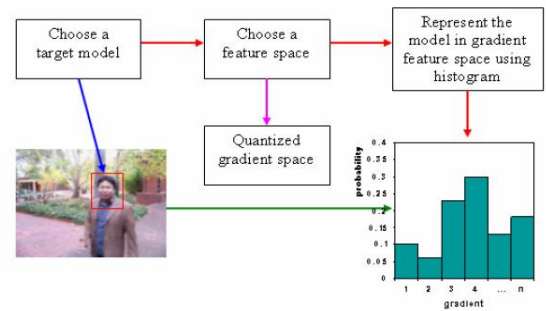

**Fig. 4.** Show target model using histogram (Zhao, 2007)

### 2.1.2 Texture

Texture is a repeated pattern of information or arrangement of the structure with regular intervals. Texture features are not obtained directly. These features are generated by image preprocessing techniques.

The texture feature is an important feature in the image, it can be used along with the color feature to describe the contents of an image or a region of the image. Because the color feature is not sufficient to identify similar objects and sometimes it can be seen that different images have the same histogram.

Gabor wavelet (Barina, 2016) is one of the most studied texture features. The most important property of Gabor filters is their invariance to illumination, rotation, scale, and translation which makes it suitable for object tracking. A method to detect the location and body shapes of moving animals using a Gabor filter is presented in (Wagenaar and Kristan, 2010). The local binary pattern is another textual descriptor (Ojala et al., 1994). Zhao et al. have used LBP to describe moving objects and also have used a Kalman filter for target tracking (Zhao et al., 2009).

Fig. 5 shows the Gabor wavelets of the face image, which contain a series of Gabor kernels (Li et al., 2017).

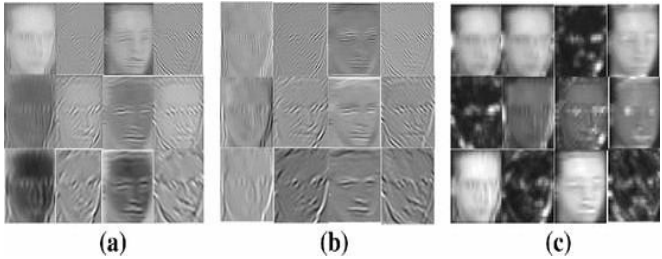

**Fig. 5.** (a): Real part of the Gabor kernels. (b): Imaginary part of the Gabor kernels. (c): Amplitude of the Gabor kernels (Li et al., 2017)

### 2.1.3 Optical Flow

Optical flow is the apparent motion of brightness patterns in the image. apparent motion can be caused by lighting changes without any actual motion. The optical flow algorithm calculates the displacement of brightness patterns from one frame to another. Algorithms that calculate displacement for all image pixels are known as dense optical streaming algorithms, while sparse algorithms estimate displacement tension for a selective number of pixels in an image (Barron et al., 1992).

In the Fig. 6 four successive frames are analyzed and optical flow is calculated for them. The green lines are the motion vectors with their end points shown in red dot. As you can see, some objects are marked with a red dot, which indicates that the object is stationary and its motion vector length is zero. The blocks which are moving more have larger motion vectors ("Optical Flow | CommonLounge," n.d.).

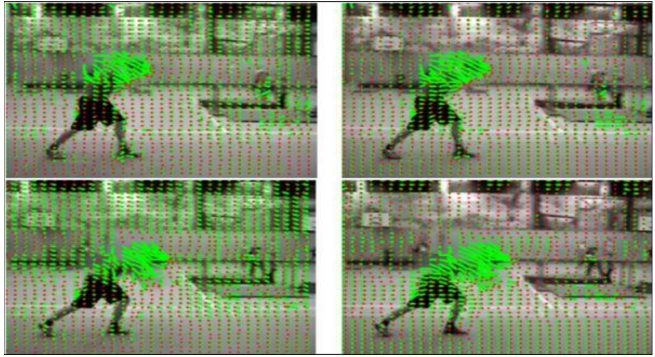

**Fig. 6.** calculate optical flow in frames (Kim and Kwon, 2016)

Some papers have used optical flow for tracking (Hariyono et al., 2014; Kim and Kwon, 2016).

Hariyono et al. (Hariyono et al., 2014) present a method based on optical flow and HOG that can detect pedestrians from inside a moving vehicle. To obtain the optical flow, two consecutive images are divided into 14×14 pixel grid cells. Then each cell in the current frame is followed to change the corresponding cell in the next frame.

### 2.2 Segmentation Based Methods

Segmenting foreground objects from a video frame is fundamental and the most critical step in visual tracking. Foreground segmentation is done to separate foreground objects from the background scene. Normally, the foreground objects are the objects that are moving in a scene. To track these objects, they are to be separated from the background scene (Verma, 2017). In the following, some of the object tracking methods based on the segmentation are examined.

### 2.2.1 Bottom-Up Based Method

In this type of tracking, there must be two separate tasks, first the foreground segmentation and then the object tracking.

The foreground segmentation uses a low-level segmentation to extract regions in all frames, and then some features are extracted from the foreground regions and tracked according to those features (Yao et al., 2019), as shown in Fig. 8.

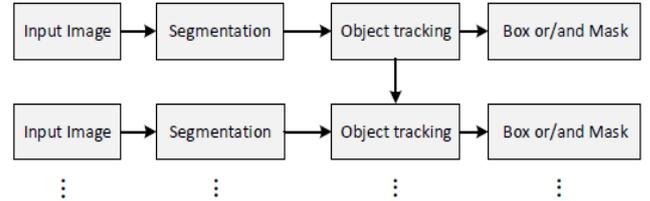

**Fig. 8.** Bottom-Up based framework (Hariyono et al., 2014)

Some papers have used a bottom-up based method for tracking (Adam et al., 2006; Cai et al., 2014; Gu and Lee, 1998; Son et al., 2015; Zhou and Tao, 2013).

Son et al. (Son et al., 2015) uses this method for object tracking that propagates posteriorly through particle filters to estimate the target state and it obtains the target area by classifying the sections of the desired area and uses the online gradient boosting decision tree to classify it.

### 2.2.2 Joint Based Method

In the bottom-up method, foreground segmentation and tracking are two separate tasks; one of the problems with this method is that the segmentation error propagates forward, causing error tracking. To solve this problem, as shown in Fig. 9, the researchers merged the foreground segmentation and tracking method, which improved tracking performance.

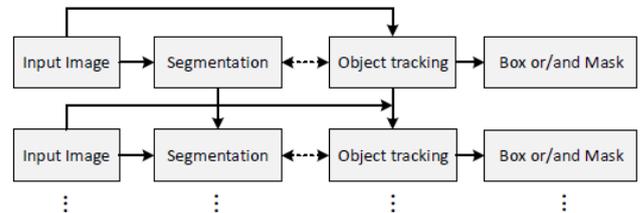

**Fig. 9.** Joint based framework (Hariyono et al., 2014)

Some papers have used joint based method for tracking (Bibby and Reid, 2008; Milan et al., 2015; Schubert et al., 2015).

Schubert et al. (Schubert et al., 2015) used this method for tracking. This paper uses a basic method that has three steps, as shown in Fig. 7. In the first step, the appearance model is constructed using a probabilistic formulation. In the second step, the model is segmented using this model; and in the third step, the tracking is performed.

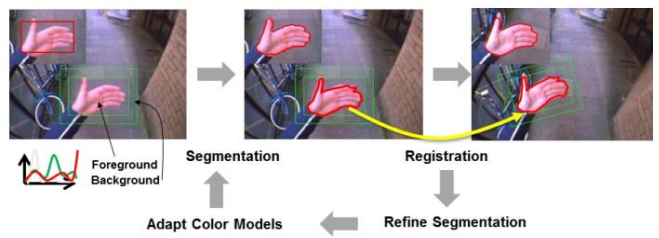

**Fig. 7.** Joint-based tracking example overview (Schubert et al., 2015)

### 2.3    Estimation Based Methods

Estimation methods formulate the tracking problem to an estimation problem in which an object is represented by a state vector. The state vector describes the dynamic behavior of a system, such as the position and velocity of an object. The general framework for the dynamic mode estimation problem is taken from Bayesian methods (Verma, 2017).

The Bayesian filters allow the target to continuously update its position on the coordinate system, based on the latest sensor data. This algorithm is recursive and consists of two steps: prediction and updating.

The prediction step estimates the new position of the target in the next step using the state model, while the updating step uses the current observation to update the target position using the observation model. The prediction and updating steps are performed on each frame of the video. Here are some examples of this method.

#### 2.3.1    Kalman Filter

To use the Kalman filter (Najafzadeh et al., 2015) in the object tracking, a dynamic model of the target movement should be designed. The Kalman filter is used to estimate the position of a linear system assumed that the errors are Gaussian. In many cases, dynamic models are nonlinear, so in this case, the Kalman filter is not used and other suitable algorithms are used. One of these algorithms is the extended Kalman Filter. The framework for using the Kalman filter is shown in Fig. 10.

Kalman filter has important features that tracking can take advantage of it, including:

• Predict the future location of the object

• Forecast correction based on new measurements

• Noise reduction caused by incorrect diagnoses

The Kalman filter consists of two stages of prediction and updating. In the prediction step, previous models are used to predict the current position. The update step uses current measurements to correct the position.

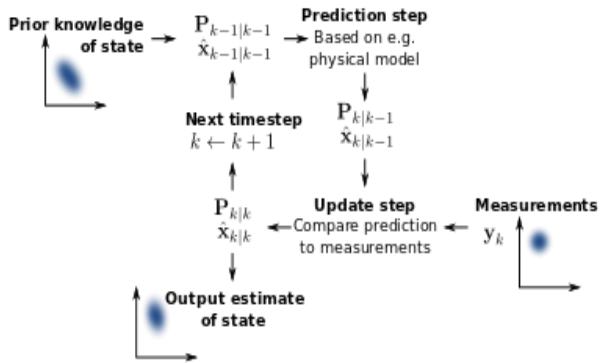

**Fig. 10.** Kalman filter framework

Some papers have used kalman filter for tracking (Gunjal et al., 2018; Taylor et al., 2021; Zhao et al., 2009).

Gunjal et al. (Gunjal et al., 2018) used the Kalman filter to object tracking, where each object can be tracked from the selected frame. In the selected frame, the object can be selected for tracking and the subsequent frames can be tracked.

#### 2.3.2    Particle Filter

Most tracking issues are non-linear. Therefore, particle filter has been considered for solving such problems. The particle filter is a recursive Monte Carlo statistical calculation method that is often used for non-gaussian noise measurement models. The main idea of the particle filter shows the distribution of a set of particles.

Each particle has a probability weight, which represents the probability of sampling of this particle based on probability density function. One of the problems with this method is that the particles that have more probability to be selected several times, and resampling is used to solve this problem. The framework for using particle filters is shown in Fig. 11.

Particle filters are also used in (Li et al., 2008; Mihaylova et al., 2007; Mondal, 2021; Pérez et al., 2002).

Mihaylova et al. (Mihaylova et al., 2007) because of the power and adaptability of Monte Carlo techniques, focus more on them. Several particles are produced, each particle having a weight that indicates the quality of a particular particle. The estimation of the desired variable is obtained from the weighted sum of the particles. It has two important steps: prediction and updating. During the prediction, each particle is modified according to the state model. At the update stage, each particle's weight is re-evaluated with new data. The method eliminates the re-sampling of light-weight particles and repeats the particles at larger weights.

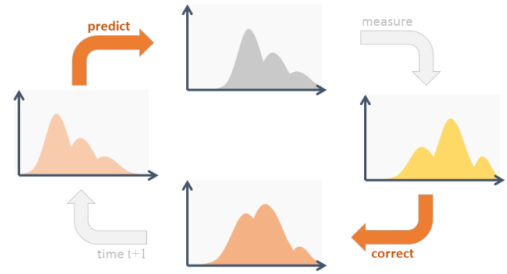

**Fig. 11.** Particle filter framework

Also, Ding et al. (Ding et al., 2016) present the object tracking based on histogram-based particle filtering. The main purpose of this method is to investigate the appearance changes, especially the state and brightness of the object. Fig. 12 shows the block diagram of this method.

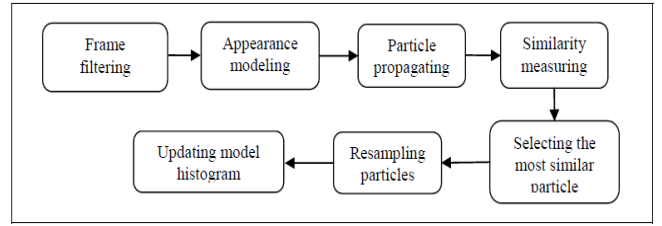

**Fig. 12.** Block diagram based on histogram based particle filter (Ding et al., 2016)

The idea behind this method is to use a FIFO queue of target histograms to overcome object brightness changes and transformations. The model histogram is calculated based on the weighted average of the histogram and is recursively calculated.

### 2.4    Learning-Based Methods

In learning-based methods, the features and appearance of different targets and their prediction are learned in the next frames, and then in the test time, based on what they learned, they can detect the object and track it in the next frames. Learning-based methods are often divided into three types of generative, discriminative and reinforcements learning.

#### 2.4.1    Discriminative Methods

Discriminative trackers usually consider tracking as a classification problem that discriminates the target from the background. Discriminative learning is divided into two categories contains Shallow learning and Deep learning.

### 2.4.1.1 Shallow Learning

Shallow learning methods is a type of methods that use pre-defined features and they can't extract features.

Object tracking can be considered as a decision-making process where classification is trained to learn the discrimination between the target and the background. After the training, at the test time, it decides that the object is target or not. This methods use pre-defined features of different objects to identify objects and then various methods such as support vector machines (Avidan, 2004; Malisiewicz et al., 2011; Tian et al., 2007; Zhang et al., 2015), random forest (Saffari et al., 2009), decision tree(Xiao et al., 2015) be used for classification.

Tian et al. (Tian et al., 2007), considered tracking as a binary classification problem, and SVM is selected as the main classification. In object tracking, as shown in Fig. 14, the target area defined as positive data, and the surrounding environment is defined as negative data. The goal is to train an SVM classifier that can classify positive and negative data into new frames. Starting with the first frame, positive and negative examples are used to learn SVM classification. Then the search area can be estimated in the next frame. Finally, the target area is selected in the next frame with the maximum local score in the search area.

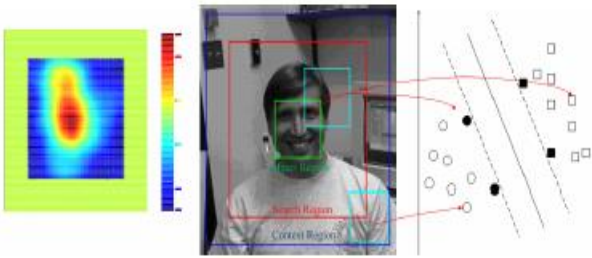

**Fig. 14.** Overview of Using Support Vector Machine in Tracking (Tian et al., 2007)

### 2.4.1.2 Deep Learning

Shallow learning with fewer layers predicts the model, but deep learning has too many layers. Another difference is that shallow learning requires important and discriminatory features extracted by professionals, but deep learning itself extracts these important features. Deep learning (LeCun et al., 2015) has made impressive developments in various areas, including computer vision, in recent years. One of the areas where deep learning has affected it and increased its accuracy is object tracking. Due to this increasing accuracy, we will focus more on deep learning methods which are explained below. In this paper, deep learning methods are classified into feature extraction based and end-to-end methods.

#### 2.4.1.2.1 Feature extraction Based Methods

Wang et al. show that feature extraction is also a very important issue in the design of a robust tracker (Wang et al., 2015). From the research experience of classical tracking algorithms, it can be concluded that any development in the methods of feature extraction or machine learning techniques may lead to the development of tracking. Therefore, since deep learning techniques have shown great abilities in feature extraction and object classification, it can be concluded that the use of deep learning can improve tracking performance. Due to the success of the deep features in image classification, some of the deep-network tracking methods are used to extract features that are known as a feature extraction network.

These methods separate the detection and tracking sections. Detection uses deep learning methods that can extract deep features, but for tracking, these methods can be classified into two categories of tracking with classical methods and tracking with Deep methods.

Yang et al have used Faster RCNN, KCF2 and Kalman filter trackers (Bu et al., n.d.). Chahyati et al. have used Faster RCNN and the Siamese network(Chahyati et al., 2017). Agarwal and Suryavanshi have used Faster RCNN and GOTURN (Agarwal and Suryavanshi, 2017). Ghaemmaghami has used SSD, YOLO, and LSTM (Ghaemmaghami, n.d.), and an example of the architecture of this method is shown in Fig. 13.

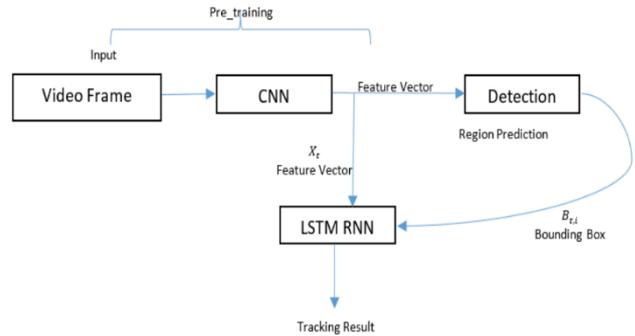

**Fig. 13.** Feature extraction based tracking example (Ghaemmaghami, n.d.)

#### 2.4.1.2.2 End-To-End Methods

End-to-end methods train a network to conduct both feature extraction and candidate evaluation. In this paper, end-to-end methods are classified into three categories of Siamese trackers, patch learning and graph-based trackers.

##### 2.4.1.2.2.1 Siamese Tracker

Siamese networks have two inputs and produce one output. It captures two inputs and measures the similarity between the two images to determine that the same objects exist in two images or not.

These types of networks are capable of learning similarities and common features. There are some papers that use the siamese tracker method for tracking such as (Chen and Tao, 2017; Held et al., 2016; Jiang et al., 2021; Tao et al., 2016; Wang et al., 2018). An example of this tracker is GOTURN tracker. This tracker proposed a method for offline training of neural networks that can detect new objects at 100 fps during the test (Held et al., 2016). This tracker is significantly fast. This tracker uses a simple feedforward neural network and does not require online training. The tracker learns the general relationship between object movement and appearance and can be used to track new objects

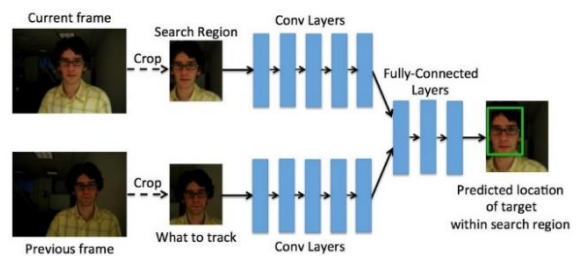

**Fig. 15.** GOTURN Tracker Architecture (Held et al., 2016)

that do not appear in the training set. The architecture of this network is shown in Fig. 15.

In this architecture, the target object and the search area are assigned to the layers of convolution. The output of the convolution layer is a set of features that is a high-level representation of the image. These features are given as inputs to the fully connected layers. The fully connected layers compare the features of the target and current object, and they somehow trained to take into account challenges such as occlusion, rotation, and so on. Finally, the last layer is fully connected to output with 4 nodes whose output is a bounding box.

### 2.4.1.2.2.2    *Patch learning tracker*

In patch learning method, positive and negative samples are extracted. Then, the method trains the neural network model on these samples. Finally, the model is tested on some selected samples, and the maximum response indicates the target position. Some papers have used patch learning methods for tracking (Han et al., 2017; Jung et al., 2018; Nam and Han, 2016; Soleimanitaleb et al., 2020). An example of this method is MDNET tracker (Nam and Han, 2016). For online training of this tracker, as can be seen in Fig. 17, 50 positive samples and 50 negative samples are drawn. A positive sample means that the IOU of the sample and the target is >= 0.7 and the negative sample <= 0.3. Then, based on these samples, the weight of the fully connected layers are corrected.

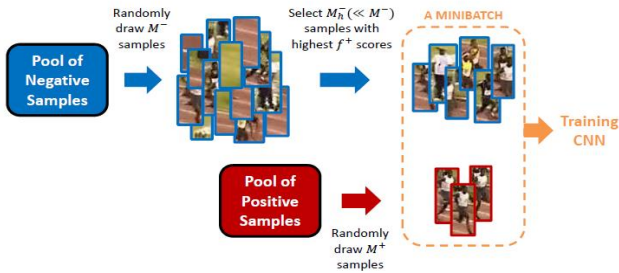

**Fig. 17.** Patch based tracking in MDNET network (Nam and Han, 2016)

### 2.4.1.2.2.3    *Graph-based tracker*

In computer vision, graph theory has been successfully applied to solve many tasks. Graphs offer a rich and compact representation of the spatial relations between objects in an image and they can be used to model many types of relations and processes.

There are some papers that use graph-based methods for tracking such as (Gomila and Meyer, 2003; Nam et al., 2016; Wang and Ling, 2017). An example of this method is TCNN tracker (Nam et al., 2016). The proposed algorithm in TCNN tracker uses convolutional neural networks for target representation, as several CNNs collaborate to estimate target states and determine the optimal paths for model updating online in the tree. Maintaining multiple CNNs in the various branches of the tree structure is suitable to counter multimode target representation and maintain model reliability through smooth updates along tree paths. Because multiple CNNs share all the parameters in the layers of the convoluted layer, it takes advantage of several models by saving memory space and preventing additional network evaluations. The final state of the target is estimated by sampling the target candidates in all cases in the previous frame and identifying the best sample in terms of the average weighted score from a set of active CNNs. Fig. 16 shows the target state estimation and the TCNN method model update methods. The width of a black arrow indicates the weight of a CNN for estimating the target state, while the width of a red edge indicates the dependence between the two CNNs. A rectangular box width means the CNN reliability associated with that box.

When a new frame is given, the candidate samples are plotted around the target position estimated in the previous frame, and the probability of each sample is calculated based on the weighted average of several CNNs. The weight of each CNN is determined by the reliability of the path that the CNN has updated to the tree structure. The target state in the current frame is estimated by finding the candidate with the maximum probability.

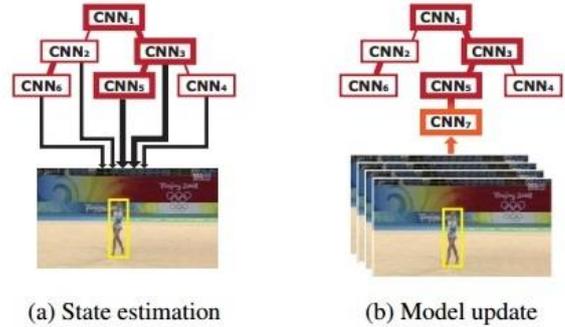

**Fig. 16.** TCNN method architecture (Nam et al., 2016)

### 2.4.2    Generative Methods

The generative appearance models mainly concentrate on how to accurately fit the data from the object class. However, it is very difficult to verify the correctness of the specified model in practice. By introducing online-update mechanisms, they incrementally learn visual representation for the foreground object region information while ignoring the influence of the background. Traditional online learning methods are adopted to track an object by searching for regions most similar to the target model. The online learning strategy is embedded in the tracking framework to update the appearance model of the target adaptively in response to appearance variations.

There are some papers that use generative methods for tracking such as (Bao et al., 2012; Jia et al., 2012).

### 2.4.3    Reinforcement Learning

In a reinforcement learning (Harmon and Harmon, 1997) problem, we encounter an agent that interacts with the environment through trial and error and learns to select the optimal action to achieve the goal.

There are some papers that use a reinforcement learning method for tracking such as (Choi et al., 2017; Valmadre et al., 2017; Yun et al., 2017).

One example of this approach is (Choi et al., 2017). As shown in Fig. 18, this network is divided into two parts; the first part is the matching network and the second part is the policy network. Assuming the networks are trained and their weights are constant, tracking can be done in the desired sequence. For a given frame, it crops a search image based on previous target bounding box information. The search image is cut from the center of the previous target location and its scale is the same as the previous target scale. Using the appearance templated that obtain from previously tracked frames, the matching network generates prediction maps for each template. Each predict map is then fed to the policy network, where it generates the scores for each predict map. The prediction map with the maximum score is selected and the corresponding pattern is used for target tracking. The position in the prediction map corresponds to the next position of the target in the search image. The patterns appear at a regular interval during the tracking process. The policy network is trained to train in a variety of situations.

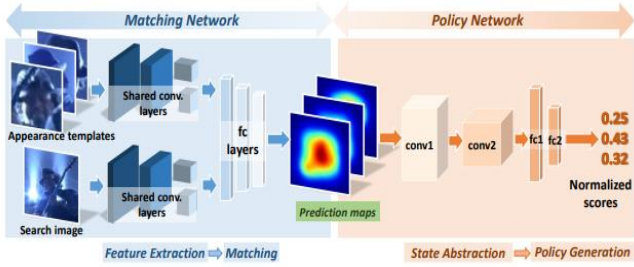

**Fig. 18.** An example of reinforcement learning object tracking (Choi et al., 2017)

## 3    Datasets

This section examines the various datasets available in object tracking. Some of the datasets and their general information are listed in Table 2, which will be discussed in detail below.

**Table 2: Object tracking datasets**

| dataset | No. of videos | NO. of classes |
|---|---|---|
| OTB100 (Wu et al., 2015) | 100 | 16 |
| OTB50 (Wu et al., 2013) | 50 | - |
| VOT2013 (Kristan et al., 2013) | 17 | - |
| VOT2014 (LIRIS, n.d.) | 25 | 11 |
| VOT2015 (Kristan et al., 2015) | 60 | - |
| VOT2017 (Kristan et al., 2017) | 60 | 24 |
| VOT2018 (Kristan et al., 2018) | 60 | - |
| VOT2019 (Kristan et al., 2019) | 60 | 30 |
| DTB70 (Li and Yeung, 2017) | 70 | 15 |
| Nfs (Kiani Galoogahi et al., 2017) | 100 | 17 |
| UAV123 (Mueller et al., 2016) | 123 | 9 |
| TempleColor128 (Liang et al., 2015) | 129 | 27 |
| ALOV300++ (Smeulders et al., 2013) | 314 | 59 |
| NUS-PRO (Li et al., 2015) | 365 | 8 |
| OxUvA (Valmadre et al., 2018) | 366 | 22 |
| LaSOT (Fan et al., 2019) | 1400 | 70 |
| GOT-10k (Huang et al., 2019) | 10000 | 563 |
| TrackingNet (Muller et al., 2018) | 31000 | 27 |
| YouTube-BoundingBoxes (Real et al., 2017) | 380000 | 23 |

### 3.1    OTB100

This dataset contains 100 video sequences. The videos are labeled with 11 attributes, namely: illumination variation, scale variation, occlusion, deformation, motion blur, fast motion, in-plane rotation, out-of-plane rotation, out-of-view, background clutters, and low resolution. An example of this dataset is shown in Fig. 19.

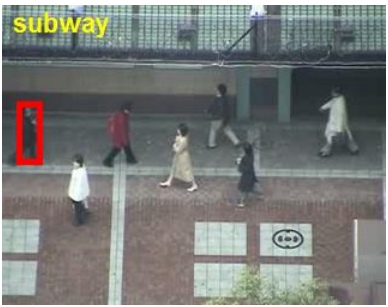

**Fig. 19.** example of OTB100 dataset (Wu et al., 2015)

### 3.2    OTB50

50 of the most challenging OTB100 movies selected and named OTB50. An example of this dataset is shown in Fig. 20.

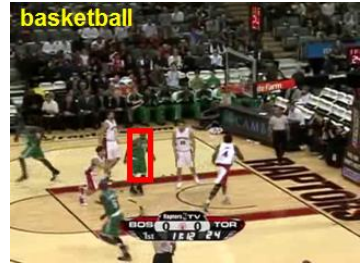

**Fig. 20.** example of OTB50 dataset (Wu et al., 2013)

### 3.3    VOT

These datasets have different versions in different years and each version has a different number of sequences. VOT2015-2019 are the only object tracking datasets that labeled with rotated bounding boxes. Each frame annotated by 6 attributes, namely: occlusion, illumination change, object motion, object size change, camera motion, unassigned. An example of this dataset is shown in Fig. 21.

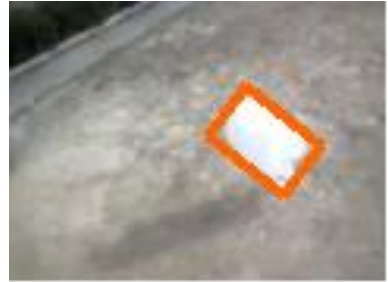

**Fig. 21.** example of VOT dataset (Kristan et al., 2015)

### 3.4    ALOV300++

The Amsterdam Library of Ordinary Videos for tracking is trying to cover a variety of conditions and challenges. Including: illuminations, transparency, specularity, confusion with similar objects, clutter, occlusion, zoom, severe shape changes, motion patterns, low contrast, and so on. ALOV300 is designed to be consistent with other benchmarks for tracking in various aspects such as light, albedo, transparency, motion smoothness, confusion, occlusion and shaking camera. The dataset consists of 315 video sequences. The main source of the data is real-life videos from YouTube with 64 different types of targets. To maximize diversity, most of the sequences are short. The total number of frames in ALOV300 is 89364. The data in ALOV300 are annotated by a rectangular bounding box along the main axes of flexible size every fifth frame. An example of this dataset is shown in Fig. 22.

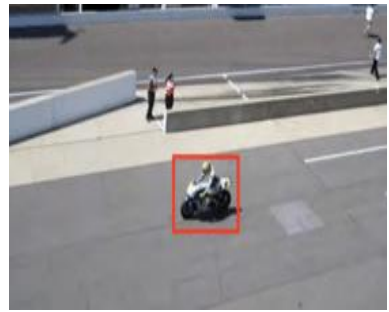

**Fig. 22.** example of ALOV300++ dataset (Smeulders et al., 2013)

### 3.5    TempleColor128

This dataset contains 128 videos that are specifically designated to evaluate color-enhanced trackers. The videos are labeled with 11 similar attributes as in OTB. An example of this dataset is shown in Fig. 23.

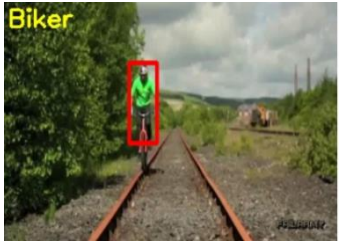

**Fig. 23.** example of TempleColor128 dataset (Liang et al., 2015)

### 3.6    NUS-PRO

This dataset contains 365 sequences that focus on human and rigid object tracking. Each sequence is annotated with both the target location and the occlusion level for evaluation. An example of this dataset is shown in Fig. 24.

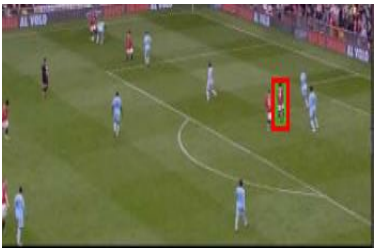

**Fig. 24**: example of NUS_PRO dataset (Li et al., 2015)

### 3.7    DTB70

The DTB3 dataset contains 70 video sequences with RGB data that are annotated ground truth bounding boxes in all video frames. Some of the videos are taken by a drone, and most are tracking people and cars. Other videos are coming together from YouTube add more diversity to both the target appearance and the scene itself.

Each subset of sequences corresponds to one of the attributes, namely: scale variation, aspect ratio variation, occlusion, deformation, fast camera motion, in-plane rotation, out-of-plane rotation, out-of-view, background cluttered, similar objects around, and motion blur. An example of this dataset is shown in Fig. 25.

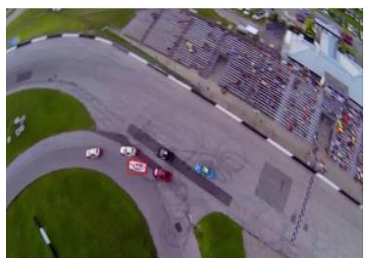

**Fig. 25.** example of DTB70 dataset (Li and Yeung, 2017)

### 3.8    Nfs

Nfs4 is the first higher frame rate video dataset and benchmark for visual object tracking. This dataset includes 100 videos of 380K frames, with 75 videos recorded using the iPhone 6 (and above) and iPad Pro and 25 videos taken from YouTube.

All videos are manually labeled with nine visual attributes, such as occlusion, illumination variation, scale variation, object deformation, fast motion, viewpoint change, out of view, background clutter, and low resolution. An example of this dataset is shown in Fig. 26.

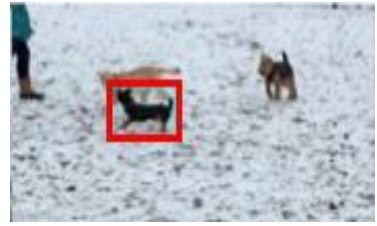

**Fig. 26.** example of Nfs dataset (Kiani Galoogahi et al., 2017)

### 3.9    UAV123

This dataset contains 123 video sequences and more than 110K frames. It was captured using cameras mounted on drones and includes all bounding boxes and attribute annotations. An example of this dataset is shown in Fig. 27.

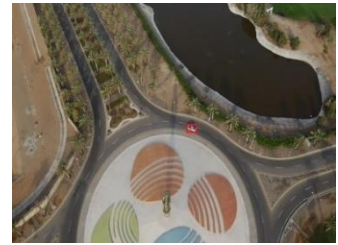

**Fig. 27.** example of UAV123 dataset (Mueller et al., 2016)

### 3.10    GOT-10k

This dataset contains over 10,000 video segments and contains 563 object classes. This database enables unified training and fair comparison of deep trackers. An example of this dataset is shown in Fig. 28.

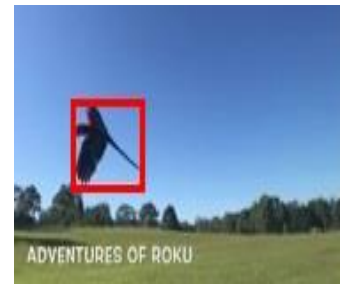

**Fig. 28.** example of GOT-10k dataset  (Huang et al., 2019)

### 3.11    LaSOT

This dataset contains 123 video sequences with 70 object categories selected from 1000 classes of ImageNet. Sequences labeled with 14 attributes, including illumination variation, full occlusion, partial occlusion, deformation, motion blur, fast motion, scale variation, camera motion, rotation, background clutter, low resolution, viewpoint change, out-of-view and aspect ratio change. An example of this dataset is shown in Fig. 29.

---

[3] Drone Tracking Benchmark

[4] Need for Speed

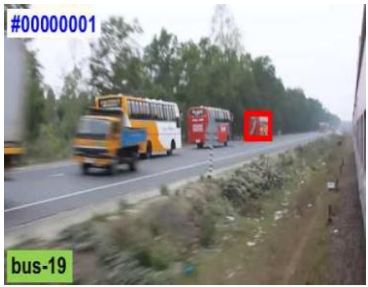

**Fig. 29.** example of LaSOT dataset (Fan et al., 2019)

### 3.12 *OxUvA*

The purpose of introducing this dataset is to collect long, realistic videos that the intended purpose disappears and reappears. This dataset is dedicated to single-object tracking, and consists of 366 videos. However, a problem with OxUvA is that it does not provide dense annotations in consecutive frames. An example of this dataset is shown in Fig. 30.

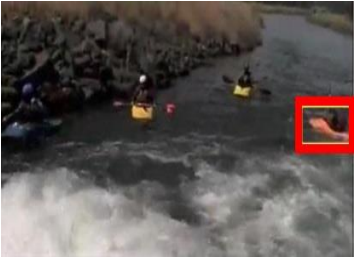

**Fig. 30.** example of OxUvA dataset (Valmadre et al., 2018)

### 3.13 *TrackingNet*

TrackingNet is a first large-scale dataset for object tracking in the wild. It includes over 30K videos with an average duration of 16.6s and more than 14M dense bounding box annotations. The dataset is not limiting to a specific context but instead covers a wide selection of object classes in a broad and diverse context. This dataset was sampled from YouTube videos and thus, represents real-world scenarios and contains a large variety of frame rates, resolutions, context and object classes.

All videos are labeled with 15 visual attributes, such as partial occlusion, full occlusion, illumination variation, scale variation, object deformation, fast motion, out of view, background clutter, low resolution, aspect ratio change, camera motion, motion blur, similar object, in-plane rotation and out-of-plane rotation. An example of this dataset is shown in Fig. 31.

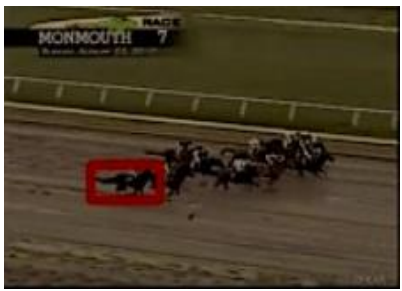

**Fig. 31.** example of TrackingNet dataset (Muller et al., 2018)

### 3.14 *YouTube-BoundingBoxes*

It is a large-scale dataset of video URLs with densely-sampled high-quality single-object bounding box annotations. This dataset contains approximately 380,000 video segments extracted from 240,000 different YouTube videos. All of these segments of the video have been annotated precisely by humans in a bounding box. An example of this dataset is shown in Fig. 32.

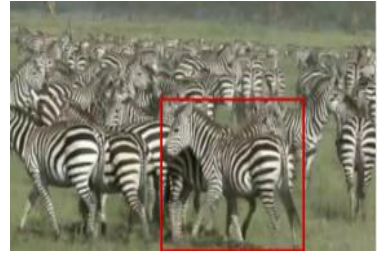

**Fig. 32.** example of YouTube-Bounding Boxes dataset (Real et al., 2017)

## 4 Evaluation Metrics

In this section, common methods for evaluating object tracking methods will be discussed.

There are several evaluation methods that have become popular in visual tracking evaluation.

Evaluating tracking algorithms with quantitative criteria is difficult because it depends on many factors, including position accuracy, tracking speed, memory required, and so on.

The criteria commonly used in tracking are:

### 4.1 *Center error*

One of the oldest methods of measuring performance in object tracking is the location error, which calculates average between the Euclidean distance of the center of the predicted target locations and the ground truth center in all frames. And because only one point is earned per frame, it's still a popular criterion.

This method has some problems, such as the center location error only measures the pixel difference and does not show the size and scale of the target. To solve this problem, the center error in each frame can be divided by the predicted target. However, despite normalization, this method can have misleading results.

In addition, when the tracker fails and the object loses its target, the error can be very large and exceed the average and not show the correct information (Čehovin et al., 2016). An example of this method is shown in Fig. 33.

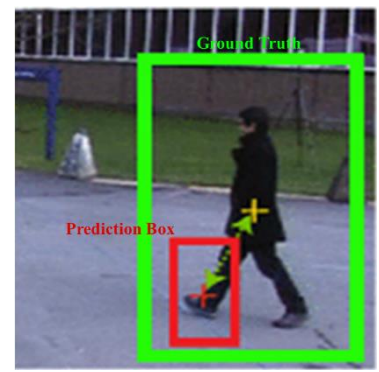

**Fig. 33.** center error example

### 4.2 Region overlap

Another common evaluation criterion is the overlap score. If $R_t^G$ is the area of the ground truth and $R_t^P$ is the area of the predicted box of frame t, the overlap score is defined as (1).

$$IOU^5 = OS = |R_t^G \cap R_t^P| / |R_t^G \cup R_t^P| \qquad (1)$$

Where t is the frame index.

An example of this method is shown in Fig. 34.

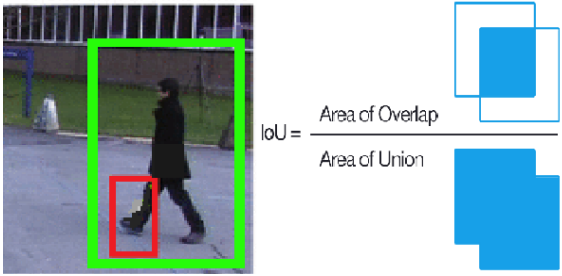

**Fig. 34.** Region overlap example

In terms of pixel classification, as shown in Fig.35, the overlap can be interpreted as (2).

$$|R_t^G \cap R_t^P| / |R_t^G \cup R_t^P| = TP/(TP + FN + FP) \qquad (2)$$

Where TP is the number of true positives, FP is the number of false positives, FN is the number of false negatives.

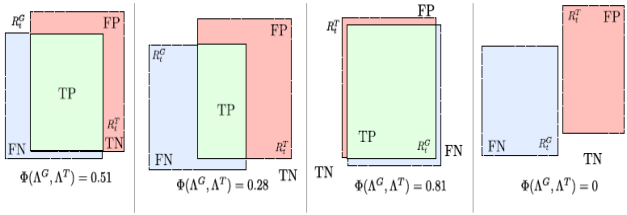

**Fig. 35.** overlap of ground-truth region with the predicted region (Čehovin et al., 2016)

Average overlap score (AOS) can be used as a performance measure. This criterion can vary from 0 to 1. The closer this number is to 1, shows that the tracker has better performance.

One of the areas overlap features is that considers both the position and the size of the bounding box predicted and ground truth at the same time, this prevents large errors from occurring when tracking failure, so this is an advantage of this method than center error. In fact, when the object tracker loses the target, the overlap score becomes zero.

In addition, as shown in Fig.36 overlap scores can be used to determine whether an algorithm successfully tracks a target in a frame, by testing whether OS is greater than a certain threshold (Čehovin et al., 2016).

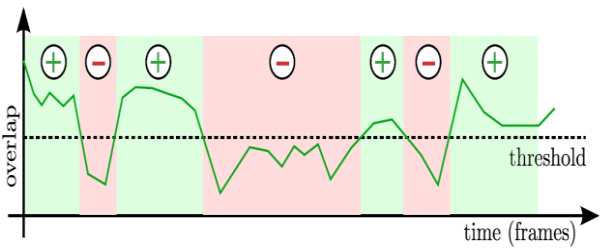

**Fig. 35.** use threshold in OS metric (Čehovin et al., 2016)

---

<sup>5</sup> Intersection over Union

### 4.3 Tracking Length

Another measurement used to compare trackers is tracking length. As shown in Fig.37 this criterion reports the number of successfully tracked frames from the beginning of the tracker to the first failure.

This criterion can be done manually, which will be biased and may have different results if repeated with the same person. To automatically determine the failure criterion can place a threshold in the center or the overlap. The choice of failure threshold may affect the comparison results, so the appropriate threshold should be chosen. But this method also has problems. One of the problems with partial review is the video sequence before failure, and the rest of the frames are left out (Čehovin et al., 2016).

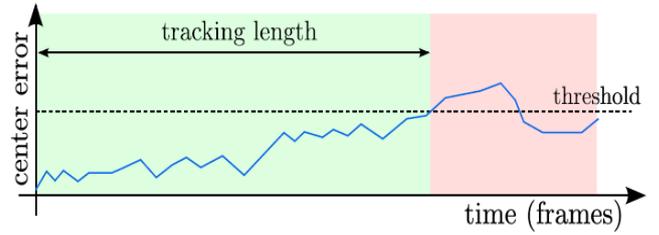

**Fig. 37.** Tracking Length performance (Čehovin et al., 2016)

### 4.4 Failure rate

The criterion that mainly solves the problems of measuring the length of the track is the measurement of the failure rate. As shown in Fig.38 failure rate considers tracking as a supervised system in which a human operator redirects the tracker after failure. The required number of manual interventions is recorded and used in each frame.

Compared to measuring tracking length, the failure rate approach has the advantage that the entire sequence is used in the evaluation process and reduces the importance of the sequence start. The problem of choosing the right failure threshold is more obvious here because any change in criteria requires a repeat experiment (Čehovin et al., 2016).

### 4.5 Performance plots

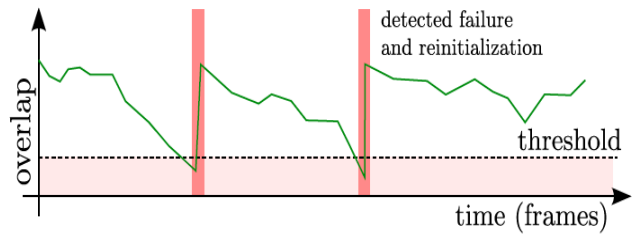

**Fig.38.** Failure rate performance (Čehovin et al., 2016)

plots are often used to visualize the behavior of a tracker when examining multiple trackers or a set of tracker parameters because it more clearly identifies tracker behavior. One of the most widely used methods of plot tracker behavior is plot based on center error, which shows the center error according to the number of frames. These types of charts are commonly used for single tracking. Another way is a plot based on region overlap.

As mentioned, determining the appropriate threshold has a significant impact on the performance of the tracker. To avoid selecting a specific threshold, the results can be presented based on the threshold measurement chart. These types of charts have similarities to the ROC curve, like monotony, intuitive visual comparison, and a similar calculation algorithm. We can use the area under the curve (AUC) to argue the performance of the trackers (Čehovin et al., 2016). An example of this method is shown in Fig. 39.

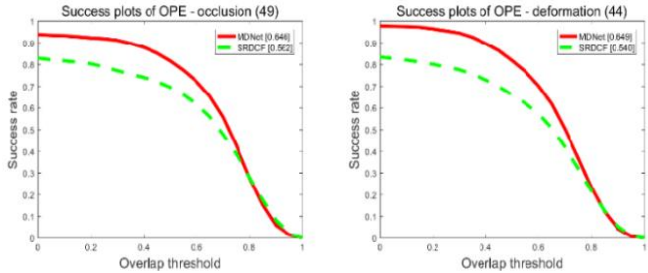

**Fig. 39.** an example of performance plots of two trackers

### 4.6    Robustness Evaluation

#### 4.6.1    one-pass evaluation (OPE)

The most common method of evaluation is to initialize an algorithm with ground-truth object mode in the first frame and report the average accuracy or success rate of all results. This simple approach is referred to as a one-pass evaluation, which has two problems. First, a tracking algorithm may be sensitive to the initial value in the first frame, and its performance may be different with different initial modes or frames. Second, most algorithms do not have re-initialization mechanisms, and tracking results do not provide significant information after tracking failure (Wu et al., 2015). The initialization of the target in the OPE method is shown in Fig. 40.

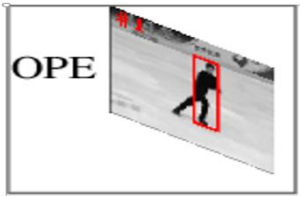

**Fig. 40.** initialize target in the first frame in OPE method

#### 4.6.2    Temporal Robustness Evaluation (TRE)

Each tracking algorithm in a sequence of images often starts from different frames. In each test, an algorithm is evaluated starting with a specific frame, with the initial value of the state-of-the-art state, until the end of an image sequence. To generate a TRE score, the tracking results are averaged from all tests (Wu et al., 2015). The initialization of the target in the TRE method is shown in Fig. 41.

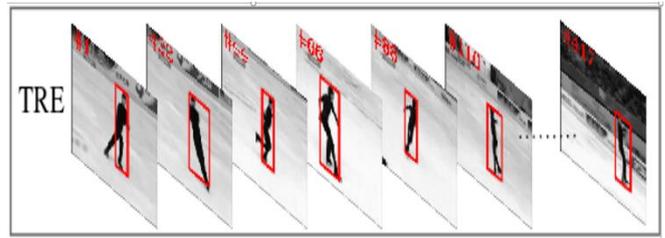

**Fig. 41.** initialize target in TRE method

#### 4.6.3    Spatial Robustness Evaluation (SRE)

Precise initial setting of a target is often important for tracking algorithms, but in the test, this is difficult t o achieve due to errors from trackers or manual tagging. To assess whether a tracking method is sensitive to initial errors, by moving or scaling the box that limits the truth of a target object, different object modes are created (Wu et al., 2015). we use 8 space shifts (4 center shifts and 4 corner shifts) and 4 scale changes. The value of the change is 10% of the target size, and the scale ratio varies from 80% to 120% of the ground truth with an incr ease of 10%.

The SRE score is the average of these 12 evaluations. The initialization of the target in the SRE method is shown in Fig. 42.

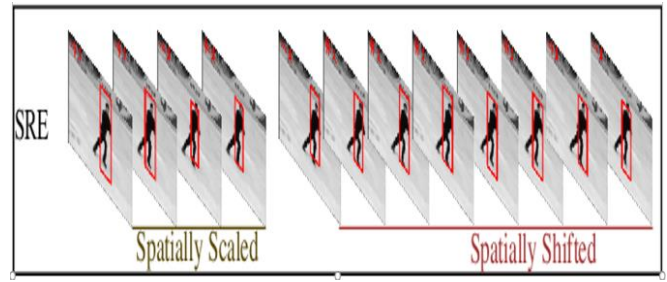

**Fig.42.** initialize target in SRE method

## 5    Conclusion

In this paper, a comprehensive classification of object tracking algorithms is presented. In this category, tracking algorithms are divided into feature-based, segmentation-based, estimation-based and learning-based categories. This paper focuses on learning-based approaches. Learning-based tracking algorithms, and especially deep learning, have recently achieved much attention. Deep learning is a new and exciting field in various fields, especially computer vision, and in many fields, it has a much higher accuracy and has made a lot of progress.

Deep learning networks have many layers and extract different features of the object in each layer. These networks have good accuracy in object tracking because it has many layers but because of model complexity are slower than shallow networks. However, it can't be said that deep learning works best in all cases and always should be used, but by knowing the advantages and disadvantages of all methods one can find out which method can work best in problem-solving.

The result of some tracker with the average overlap score criterion and on the GOT-10K, OTB100 and TrackingNet datasets shown on Table 3.

**Table 3: Tracker result with AOS criterion**

| Tracker | category | Got-10k | OTB100 | TrackingNet |
|---|---|---|---|---|
| DSST (Danelljan et al., 2014) | HOG | 0.317 | 0.470 | 0.464 |
| KCF (Henriques et al., 2014) | CF[6] | 0.279 | 0.477 | 0.447 |
| ECO (Danelljan et al., 2017) | CF | 0.395 | 0.687 | 0.554 |
| GOTURN (Held et al., 2016) | DL[7] | 0.418 | 0.450 | - |
| MDNET (Nam and Han, 2016) | DL | 0.352 | 0.660 | 0.606 |
| SiamFC (Bertinetto et al., 2016) | DL | 0.392 | 0.569 | 0.571 |

As can be seen in the Table 3, Deep Learning's methods have performed better, but depending on the dataset and evaluation criteria, the tracker's performance will be different.

## References

[1] Adam, A., Rivlin, E., Shimshoni, I., 2006. Robust fragments-based tracking using the integral histogram, in: 2006 IEEE Computer Society Conference on Computer Vision and Pattern Recognition (CVPR'06). IEEE, pp. 798–805.

[2] Agarwal, A., Suryavanshi, S., 2017. Real-Time* Multiple Object Tracking (MOT) for Autonomous Navigation. Technical report.

[3] Avidan, S., 2004. Support vector tracking. IEEE transactions on pattern analysis and machine intelligence 26, 1064–1072.

[4] Bao, C., Wu, Y., Ling, H., Ji, H., 2012. Real time robust l1 tracker using accelerated proximal gradient approach, in: 2012 IEEE Conference on Computer Vision and Pattern Recognition. IEEE, pp. 1830–1837.

[5] Barina, D., 2016. Gabor Wavelets in Image Processing. arXiv:1602.03308 [cs].

[6] Barron, J.L., Fleet, D.J., Beauchemin, S.S., Burkitt, T.A., 1992. Performance of optical flow techniques, in: Proceedings 1992 IEEE Computer Society Conference on Computer Vision and Pattern Recognition. IEEE, pp. 236–242.

[7] Bertinetto, L., Valmadre, J., Henriques, J.F., Vedaldi, A., Torr, P.H., 2016. Fully-convolutional siamese networks for object tracking, in: European Conference on Computer Vision. Springer, pp. 850–865.

[8] Bibby, C., Reid, I., 2008. Robust real-time visual tracking using pixel-wise posteriors, in: European Conference on Computer Vision. Springer, pp. 831–844.

[9] Bodor, R., Jackson, B., Papanikolopoulos, N., 2003. Vision-based human tracking and activity recognition, in: Proc. of the 11th Mediterranean Conf. on Control and Automation.

[10] Brown, M., Funke, J., Erlien, S., Gerdes, J.C., 2017. Safe driving envelopes for path tracking in autonomous vehicles. Control Engineering Practice 61, 307–316.

[11] Bu, F., Cai, Y., Yang, Y., n.d. Multiple Object Tracking Based on Faster-RCNN Detector and KCF Tracker.

[12] Cai, Z., Wen, L., Lei, Z., Vasconcelos, N., Li, S.Z., 2014. Robust deformable and occluded object tracking with dynamic graph. IEEE Transactions on Image Processing 23, 5497–5509.

[13] Čehovin, L., Leonardis, A., Kristan, M., 2016. Visual object tracking performance measures revisited. IEEE Transactions on Image Processing 25, 1261–1274.

[14] Chahyati, D., Fanany, M.I., Arymurthy, A.M., 2017. Tracking People by Detection Using CNN Features. Procedia Computer Science 124, 167–172.

[15] Chen, K., Tao, W., 2017. Once for all: a two-flow convolutional neural network for visual tracking. IEEE Transactions on Circuits and Systems for Video Technology 28, 3377–3386.

[16] Choi, J., Kwon, J., Lee, K.M., 2017. Visual tracking by reinforced decision making. arXiv preprint arXiv:1702.06291.

[17] Colchester, A.C.F., Hawkes, D.J. (Eds.), 1991. Information Processing in Medical Imaging: 12th International Conference, IPMI '91, Wye, UK, July 7-12, 1991. Proceedings, Lecture Notes in Computer Science. Springer-Verlag, Berlin Heidelberg.

[18] Danelljan, M., Bhat, G., Khan, F.S., Felsberg, M., 2017. ECO: Efficient Convolution Operators for Tracking., in: CVPR. p. 3.

[19] Danelljan, M., Häger, G., Khan, F., Felsberg, M., 2014. Accurate scale estimation for robust visual tracking, in: British Machine Vision Conference, Nottingham, September 1-5, 2014. BMVA Press.

[20] Ding, D., Jiang, Z., Liu, C., 2016. Object tracking algorithm based on particle filter with color and texture feature, in: 2016 35th Chinese Control Conference (CCC). Presented at the 2016 35th Chinese Control Conference (CCC), pp. 4031–4036. https://doi.org/10.1109/ChiCC.2016.7553983

[21] Fan, H., Lin, L., Yang, F., Chu, P., Deng, G., Yu, S., Bai, H., Xu, Y., Liao, C., Ling, H., 2019. Lasot: A high-quality benchmark for large-scale single object tracking, in: Proceedings of the IEEE Conference on Computer Vision and Pattern Recognition. pp. 5374–5383.

[22] Fiaz, M., Mahmood, A., Jung, S.K., 2018. Tracking Noisy Targets: A Review of Recent Object Tracking Approaches. arXiv:1802.03098 [cs].

[23] Fotouhi, M., Gholami, A.R., Kasaei, S., 2011. Particle filter-based object tracking using adaptive histogram, in: 2011 7th Iranian Conference on Machine Vision and Image Processing. IEEE, pp. 1–5.

[24] Ghaemmaghami, M.P., n.d. Tracking of Humans in Video Stream Using LSTM Recurrent Neural Net- work 50.

[25] Gomila, C., Meyer, F., 2003. Graph-based object tracking, in: Proceedings 2003 International Conference on Image Processing (Cat. No.03CH37429). Presented at the International Conference on Image Processing, IEEE, Barcelona, Spain, p. II-41–4. https://doi.org/10.1109/ICIP.2003.1246611

[26] Gu, C., Lee, M.-C., 1998. Semiautomatic segmentation and tracking of semantic video objects. IEEE Transactions on Circuits and Systems for Video Technology 8, 572–584.

[27] Gunjal, P.R., Gunjal, B.R., Shinde, H.A., Vanam, S.M., Aher, S.S., 2018. Moving Object Tracking Using Kalman Filter, in: 2018 International Conference On Advances in Communication and Computing Technology (ICACCT). IEEE, pp. 544–547.

[28] Han, B., Sim, J., Adam, H., 2017. Branchout: Regularization for online ensemble tracking with convolutional neural networks, in: Proceedings of the IEEE Conference on Computer Vision and Pattern Recognition. pp. 3356–3365.

[29] Hariyono, J., Hoang, V.-D., Jo, K.-H., 2014. Moving object localization using optical flow for pedestrian detection from a moving vehicle. The Scientific World Journal 2014.

[30] Harmon, M.E., Harmon, S.S., 1997. Reinforcement Learning: A Tutorial. WRIGHT LAB WRIGHT-PATTERSON AFB OH.

---

[6] Correlation Filter

[7] Deep Learning

[31] Held, D., Thrun, S., Savarese, S., 2016. Learning to track at 100 fps with deep regression networks, in: European Conference on Computer Vision. Springer, pp. 749–765.

[32] Henriques, J.F., Caseiro, R., Martins, P., Batista, J., 2014. High-speed tracking with kernelized correlation filters. IEEE transactions on pattern analysis and machine intelligence 37, 583–596.

[33] Huang, L., Zhao, X., Huang, K., 2019. Got-10k: A large high-diversity benchmark for generic object tracking in the wild. IEEE Transactions on Pattern Analysis and Machine Intelligence.

[34] Jia, X., Lu, H., Yang, M.-H., 2012. Visual tracking via adaptive structural local sparse appearance model, in: 2012 IEEE Conference on Computer Vision and Pattern Recognition. IEEE, pp. 1822–1829.

[35] Jiang, S., Xu, B., Zhao, J., Shen, F., 2021. Faster and Simpler Siamese Network for Single Object Tracking. arXiv:2105.03049 [cs].

[36] Jung, I., Son, J., Baek, M., Han, B., 2018. Real-time mdnet, in: Proceedings of the European Conference on Computer Vision (ECCV). pp. 83–98.

[37] Kiani Galoogahi, H., Fagg, A., Huang, C., Ramanan, D., Lucey, S., 2017. Need for speed: A benchmark for higher frame rate object tracking, in: Proceedings of the IEEE International Conference on Computer Vision. pp. 1125–1134.

[38] Kim, D.-S., Kwon, J., 2016. Moving object detection on a vehicle mounted back-up camera. Sensors 16, 23.

[39] Kristan, M., Leonardis, A., Matas, J., Felsberg, M., Pflugfelder, R., ˇCehovin Zajc, L., Vojir, T., Bhat, G., Lukezic, A., Eldesokey, A., 2018. The sixth visual object tracking vot2018 challenge results, in: Proceedings of the European Conference on Computer Vision (ECCV). pp. 0–0.

[40] Kristan, M., Leonardis, A., Matas, J., Felsberg, M., Pflugfelder, R., Cehovin Zajc, L., Vojir, T., Hager, G., Lukezic, A., Eldesokey, A., Fernandez, G., 2017. The Visual Object Tracking VOT2017 Challenge Results. Presented at the Proceedings of the IEEE International Conference on Computer Vision, pp. 1949–1972.

[41] Kristan, M., Matas, J., Leonardis, A., Felsberg, M., Cehovin, L., Fernandez, G., Vojir, T., Hager, G., Nebehay, G., Pflugfelder, R., 2015. The visual object tracking vot2015 challenge results, in: Proceedings of the IEEE International Conference on Computer Vision Workshops. pp. 1–23.

[42] Kristan, M., Matas, J., Leonardis, A., Felsberg, M., Pflugfelder, R., Kamarainen, J.-K., Cehovin Zajc, L., Drbohlav, O., Lukezic, A., Berg, A., 2019. The seventh visual object tracking vot2019 challenge results, in: Proceedings of the IEEE International Conference on Computer Vision Workshops. pp. 0–0.

[43] Kristan, M., Pflugfelder, R., Leonardis, A., Matas, J., Porikli, F., Cehovin, L., Vojir, T., 2013. The Visual Object Tracking VOT2013 challenge results. ICCV2013 Workshops, in: Workshop on Visual Object Tracking Challenge.

[44] Laurense, V.A., Goh, J.Y., Gerdes, J.C., 2017. Path-tracking for autonomous vehicles at the limit of friction, in: 2017 American Control Conference (ACC). IEEE, pp. 5586–5591.

[45] LeCun, Y., Bengio, Y., Hinton, G., 2015. Deep learning. Nature 521, 436–444. https://doi.org/10.1038/nature14539

[46] Li, A., Lin, M., Wu, Y., Yang, M.-H., Yan, S., 2015. Nus-pro: A new visual tracking challenge. IEEE transactions on pattern analysis and machine intelligence 38, 335–349.

[47] Li, C., Wei, W., Li, J., Song, W., 2017. A cloud-based monitoring system via face recognition using Gabor and CS-LBP features. The Journal of Supercomputing 73, 1532–1546.

[48] Li, P., Wang, D., Wang, L., Lu, H., 2018. Deep visual tracking: Review and experimental comparison. Pattern Recognition 76, 323–338.

[49] Li, S., Yeung, D.-Y., 2017. Visual object tracking for unmanned aerial vehicles: A benchmark and new motion models, in: Thirty-First AAAI Conference on Artificial Intelligence.

[50] Li, X., Zheng, N., 2004. Adaptive target color model updating for visual tracking using particle filter, in: 2004 IEEE International Conference on Systems, Man and Cybernetics (IEEE Cat. No. 04CH37583). IEEE, pp. 3105–3109.

[51] Li, Y., Ai, H., Yamashita, T., Lao, S., Kawade, M., 2008. Tracking in low frame rate video: A cascade particle filter with discriminative observers of different life spans. IEEE Transactions on Pattern Analysis and Machine Intelligence 30, 1728–1740.

[52] Liang, P., Blasch, E., Ling, H., 2015. Encoding color information for visual tracking: Algorithms and benchmark. IEEE Transactions on Image Processing 24, 5630–5644.

[53] LIRIS, F., n.d. The Visual Object Tracking VOT2014 challenge results.

[54] Malisiewicz, T., Gupta, A., Efros, A.A., 2011. Ensemble of exemplar-SVMs for object detection and beyond., in: Iccv. Citeseer, p. 6.

[55] Menze, M., Geiger, A., 2015. Object scene flow for autonomous vehicles, in: Proceedings of the IEEE Conference on Computer Vision and Pattern Recognition. pp. 3061–3070.

[56] Mihaylova, L., Brasnett, P., Canagarajah, N., Bull, D., 2007. Object tracking by particle filtering techniques in video sequences. Advances and challenges in multisensor data and information processing 8, 260–268.

[57] Milan, A., Leal-Taixé, L., Schindler, K., Reid, I., 2015. Joint tracking and segmentation of multiple targets, in: Proceedings of the IEEE Conference on Computer Vision and Pattern Recognition. pp. 5397–5406.

[58] Mondal, A., 2021. Occluded object tracking using object-background prototypes and particle filter. Appl Intell 51, 5259–5279. https://doi.org/10.1007/s10489-020-02047-x

[59] Mueller, M., Smith, N., Ghanem, B., 2016. A benchmark and simulator for uav tracking, in: European Conference on Computer Vision. Springer, pp. 445–461.

[60] Muller, M., Bibi, A., Giancola, S., Alsubaihi, S., Ghanem, B., 2018. Trackingnet: A large-scale dataset and benchmark for object tracking in the wild, in: Proceedings of the European Conference on Computer Vision (ECCV). pp. 300–317.

[61] Najafzadeh, N., Fotouhi, M., Kasaei, S., 2015. Object tracking using Kalman filter with adaptive sampled histogram, in: 2015 23rd Iranian Conference on Electrical Engineering. Presented at the 2015 23rd Iranian Conference on Electrical Engineering, pp. 781–786. https://doi.org/10.1109/IranianCEE.2015.7146319

[62] Nam, H., Baek, M., Han, B., 2016. Modeling and propagating cnns in a tree structure for visual tracking. arXiv preprint arXiv:1608.07242.

[63] Nam, H., Han, B., 2016. Learning multi-domain convolutional neural networks for visual tracking, in: Proceedings of the IEEE Conference on Computer Vision and Pattern Recognition. pp. 4293–4302.

[64] Ojala, T., Pietikainen, M., Harwood, D., 1994. Performance evaluation of texture measures with classification based on Kullback discrimination of distributions, in: Proceedings of 12th International Conference on Pattern Recognition. IEEE, pp. 582–585.

[65] Optical Flow | CommonLounge [WWW Document], n.d. URL https://www.commonlounge.com/discussion/1c2eaa85265f47a3a0a8ff1ac5fbce51 (accessed 10.17.20).

[66] Pérez, P., Hue, C., Vermaak, J., Gangnet, M., 2002. Color-based probabilistic tracking, in: European Conference on Computer Vision. Springer, pp. 661–675.

[67] Real, E., Shlens, J., Mazzocchi, S., Pan, X., Vanhoucke, V., 2017. Youtube-boundingboxes: A large high-precision human-annotated data set for object detection in video, in: Proceedings of the IEEE Conference on Computer Vision and Pattern Recognition. pp. 5296–5305.

[68] Saffari, A., Leistner, C., Santner, J., Godec, M., Bischof, H., 2009. On-line random forests, in: Computer Vision Workshops (ICCV Workshops), 2009 IEEE 12th International Conference On. IEEE, pp. 1393–1400.

[69] Sakagami, Y., Watanabe, R., Aoyama, C., Matsunaga, S., Higaki, N., Fujimura, K., 2002. The intelligent ASIMO: System overview and integration, in: IEEE/RSJ International

Conference on Intelligent Robots and Systems. IEEE, pp. 2478–2483.

[70] Schubert, F., Casaburo, D., Dickmanns, D., Belagiannis, V., 2015. Revisiting robust visual tracking using pixel-wise posteriors, in: International Conference on Computer Vision Systems. Springer, pp. 275–288.

[71] Smeulders, A.W., Chu, D.M., Cucchiara, R., Calderara, S., Dehghan, A., Shah, M., 2013. Visual tracking: An experimental survey. IEEE transactions on pattern analysis and machine intelligence 36, 1442–1468.

[72] Soleimanitaleb, Z., Keyvanrad, M.A., Jafari, A., 2020. Improved MDNET Tracker in Better Localization Accuracy. Presented at the 2020 10th International Conference on Computer and Knowledge Engineering (ICCKE)., IEEE.

[73] Soleimanitaleb, Z., Keyvanrad, M.A., Jafari, A., 2019. Object Tracking Methods:A Review, in: 2019 9th International Conference on Computer and Knowledge Engineering (ICCKE). Presented at the 2019 9th International Conference on Computer and Knowledge Engineering (ICCKE), pp. 282–288. https://doi.org/10.1109/ICCKE48569.2019.8964761

[74] Son, J., Jung, I., Park, K., Han, B., 2015. Tracking-by-segmentation with online gradient boosting decision tree, in: Proceedings of the IEEE International Conference on Computer Vision. pp. 3056–3064.

[75] Stauffer, C., Grimson, W.E.L., 2000. Learning patterns of activity using real-time tracking. IEEE Transactions on Pattern Analysis and Machine Intelligence 22, 747–757. https://doi.org/10.1109/34.868677

[76] Tai, J.-C., Tseng, S.-T., Lin, C.-P., Song, K.-T., 2004. Real-time image tracking for automatic traffic monitoring and enforcement applications. Image and Vision Computing 22, 485–501.

[77] Tao, R., Gavves, E., Smeulders, A.W., 2016. Siamese instance search for tracking, in: Proceedings of the IEEE Conference on Computer Vision and Pattern Recognition. pp. 1420–1429.

[78] Taylor, L.E., Mirdanies, M., Saputra, R.P., 2021. Optimized object tracking technique using Kalman filter. arXiv preprint arXiv:2103.05467.

[79] Tian, B., Yao, Q., Gu, Y., Wang, K., Li, Y., 2011. Video processing techniques for traffic flow monitoring: A survey, in: 2011 14th International IEEE Conference on Intelligent Transportation Systems (ITSC). IEEE, pp. 1103–1108.

[80] Tian, M., Zhang, W., Liu, F., 2007. On-line ensemble SVM for robust object tracking, in: Asian Conference on Computer Vision. Springer, pp. 355–364.

[81] Tsai, M.-H., Pan, C.-S., Wang, C.-W., Chen, J.-M., Kuo, C.-B., 2018. RFID Medical Equipment Tracking System Based on a Location-Based Service Technique. Journal of Medical and Biological Engineering 39. https://doi.org/10.1007/s40846-018-0446-2

[82] Valmadre, J., Bertinetto, L., Henriques, J., Vedaldi, A., Torr, P.H.S., 2017. End-To-End Representation Learning for Correlation Filter Based Tracking. Presented at the Proceedings of the IEEE Conference on Computer Vision and Pattern Recognition, pp. 2805–2813.

[83] Valmadre, J., Bertinetto, L., Henriques, J.F., Tao, R., Vedaldi, A., Smeulders, A.W., Torr, P.H., Gavves, E., 2018. Long-term tracking in the wild: A benchmark, in: Proceedings of the European Conference on Computer Vision (ECCV). pp. 670–685.

[84] Verma, R., 2017. A Review of Object Detection and Tracking Methods. International Journal of Advance Engineering and Research Development 4, 569–578.

[85] Wagenaar, D.A., Kristan, W.B., 2010. Automated video analysis of animal movements using Gabor orientation filters. Neuroinformatics 8, 33–42.

[86] Wang, N., Shi, J., Yeung, D.-Y., Jia, J., 2015. Understanding and diagnosing visual tracking systems, in: Proceedings of the IEEE International Conference on Computer Vision. pp. 3101–3109.

[87] Wang, T., Ling, H., 2017. Gracker: A graph-based planar object tracker. IEEE transactions on pattern analysis and machine intelligence 40, 1494–1501.

[88] Wang, X., Li, C., Luo, B., Tang, J., 2018. Sint++: robust visual tracking via adversarial positive instance generation, in: Proceedings of the IEEE Conference on Computer Vision and Pattern Recognition. pp. 4864–4873.

[89] Wu, Y., Lim, J., Yang, M.-H., 2015. Object tracking benchmark. IEEE Transactions on Pattern Analysis and Machine Intelligence 37, 1834–1848.

[90] Wu, Y., Lim, J., Yang, M.-H., 2013. Online object tracking: A benchmark, in: Proceedings of the IEEE Conference on Computer Vision and Pattern Recognition. pp. 2411–2418.

[91] Xiao, J., Stolkin, R., Leonardis, A., 2015. Single target tracking using adaptive clustered decision trees and dynamic multi-level appearance models, in: 2015 IEEE Conference on Computer Vision and Pattern Recognition (CVPR). Presented at the 2015 IEEE Conference on Computer Vision and Pattern Recognition (CVPR), pp. 4978–4987. https://doi.org/10.1109/CVPR.2015.7299132

[92] Yao, R., Lin, G., Xia, S., Zhao, J., Zhou, Y., 2019. Video Object Segmentation and Tracking: A Survey. arXiv preprint arXiv:1904.09172.

[93] Yun, S., Choi, J., Yoo, Y., Yun, K., Young Choi, J., 2017. Action-decision networks for visual tracking with deep reinforcement learning, in: Proceedings of the IEEE Conference on Computer Vision and Pattern Recognition. pp. 2711–2720.

[94] Zhang, S., Yu, X., Sui, Y., Zhao, S., Zhang, L., 2015. Object tracking with multi-view support vector machines. IEEE Transactions on Multimedia 17, 265–278.

[95] Zhao, A., 2007. Robust histogram-based object tracking in image sequences, in: 9th Biennial Conference of the Australian Pattern Recognition Society on Digital Image Computing Techniques and Applications (DICTA 2007). IEEE, pp. 45–52.

[96] Zhao, Z., Yu, S., Wu, X., Wang, C., Xu, Y., 2009. A multi-target tracking algorithm using texture for real-time surveillance, in: 2008 IEEE International Conference on Robotics and Biomimetics. IEEE, pp. 2150–2155.

[97] Zhou, T., Tao, D., 2013. Shifted subspaces tracking on sparse outlier for motion segmentation, in: Twenty-Third International Joint Conference on Artificial Intelligence. Citeseer.